\documentclass{article} %
\usepackage{collas2022_conference,times}

\usepackage{amsmath,amsfonts,bm}

\def\eqref#1{equation~\ref{#1}}

\def\1{\bm{1}}

\DeclareMathAlphabet{\mathsfit}{\encodingdefault}{\sfdefault}{m}{sl}
\SetMathAlphabet{\mathsfit}{bold}{\encodingdefault}{\sfdefault}{bx}{n}

\newcommand{\R}{\mathbb{R}}

\collasfinalcopy

\usepackage{hyperref}
\hypersetup{
    colorlinks=true,
    linkcolor=red,
    filecolor=magenta,      
    urlcolor=blue,
    citecolor=purple,
    pdfpagemode=FullScreen,
    }

\usepackage{floatrow}
\usepackage{hypcap}
\usepackage{enumitem}
\usepackage{amsmath}
\usepackage{amssymb}
\usepackage{algorithm,algorithmicx}
\usepackage[noend]{algpseudocode}
\usepackage{booktabs, multirow}

\newcommand{\mb}{\mathbf}
\newcommand{\mcal}[1]{\mathcal{#1}}
\usepackage{floatrow}
\newfloatcommand{capbtabbox}{table}[][\FBwidth]

\newcommand{\opt}{O}
\newcommand{\e}{\mb e}

\newcommand{\oihrltext}{Option-Indexed Hierarchical Reinforcement Learning}
\newcommand{\oihrl}{{\sc OI-HRL}}
\newcommand{\primitives}{base tasks}
\usepackage{graphicx}

\usepackage{wrapfig}
\usepackage{subcaption}

\title{Matching options to tasks using Option-Indexed Hierarchical Reinforcement Learning}

\author{Kushal Chauhan\textsuperscript{\normalfont 1}, Soumya Chatterjee\textsuperscript{\normalfont 1}, Akash Reddy\textsuperscript{\normalfont 2}, Balaraman Ravindran\textsuperscript{\normalfont 2}, Pradeep Shenoy\textsuperscript{\normalfont 1}\\
\textsuperscript{1}Google Research India, \textsuperscript{2}Indian Institute of Technology Madras\\
\texttt{\{kushalchauhan, soumyach, shenoypradeep\}@google.com}\\ \texttt{ee17b001@smail.iitm.ac.in}, \texttt{ravi@cse.iitm.ac.in}}

\begin{document}

\maketitle

\begin{abstract}
   The options framework in Hierarchical Reinforcement Learning breaks down overall goals into a combination of options or simpler tasks and associated policies, allowing for abstraction in the action space. Ideally, these options can be reused across different higher-level goals; indeed, such reuse is necessary to realize the vision of a continual learning agent that can effectively leverage its prior experience.  Previous approaches have only proposed limited forms of transfer of prelearned options to new task settings. We propose a novel \textit{option indexing} approach to hierarchical learning (\oihrl), where we learn an affinity function between options and the items present in the environment. This allows us to effectively reuse a large library of pretrained options, in zero-shot generalization at test time, by restricting goal-directed learning to only those options relevant to the task at hand. We develop a meta-training loop that learns the representations of options and environments over a series of HRL problems, by incorporating feedback about the relevance of retrieved options to the higher-level goal.  We evaluate \oihrl\ in two simulated settings -- the CraftWorld and AI2THOR environments -- and show that  we achieve performance competitive with oracular baselines, and substantial gains over a baseline that has the entire option pool available for learning the hierarchical policy. 
\end{abstract}

\section{Introduction}

Hierarchical reinforcement learning (HRL) splits the problem of learning a complex, multi-step task into simpler, smaller \textit{subtasks} (referred to as \textit{options} in the options framework~\citep{sutton1999between} for HRL), and a higher level policy that can put together these subtasks in different combinations to solve the overall task. We propose a setting that has not received much attention in the literature: efficient retrieval of \textit{relevant options} for a new HRL task, from a large library of prelearned options. This capability is necessary for a continual learning agent solving multiple related tasks sequentially in a given domain. In sparse reward settings, it is critical for the hierarchical policy to only explore necessary parts of the search space; even including a small number of irrelevant options can significantly impact convergence rates, or even task completion. In sharp contrast, existing option reuse approaches do not solve this retrieval problem~\citep{florensa2017stochastic,heess2016learning,hausman2018learning,shu2017hierarchical} as they only address simpler settings where a small set of options, all relevant, are reused or fine-tuned in very closely related tasks. The primary challenge here is to determine the relevant options for a particular task without first solving the task itself. 

Our key insight is that the state of the environment, and the actions enabled by items in the environment~\citep{gibson1977theory, affordances}, provides substantial information about what tasks are \textit{achievable}.
Based on this insight, we propose \oihrltext\ (\oihrl), which solves new goals in a known domain by first fetching the relevant options from the option library, then constructing a hierarchical policy using the fetched options.  Our proposal has two major ideas:  (1) an affinity score between a given environment state and the options relevant to goals achievable in that environment, and (2) a meta-training loop, which learns this affinity score by solving a series of related HRL tasks in a given problem domain. We describe these two components in Section~\ref{sec:methods}, including the characteristics of the problem domain that enable our proposal. We show in our experiments (Section~\ref{sec:results}) that \oihrl\ is substantially more effective than learning a HRL policy over the entire library of options. Further, we show that \oihrl\ is highly competitive relative to an oracle that selects exactly the option set required for the current goal, and much better than a ``noisy oracle'' which fetches a fixed number of additional irrelevant options. We conclude with a discussion of the broader implications of our findings, and potential next steps.

\section{Related Work} 
{ \bf Option transfer/reuse}
Recent works have proposed many HRL methods for transferring learned skills to solve a variety of diverse tasks. To ensure effective transfer, the skills are either pre-trained on handcrafted simple tasks~\citep{heess2016learning}, proxy rewards~\citep{florensa2017stochastic}, maximizing diversity~\citep{eysenbach2018diversity}, or obtained by training shared skills on multiple related tasks~\citep{igl2020multitask,frans2017meta}. When encountering a new task, the skills are either frozen and reused~\citep{florensa2017stochastic,heess2016learning,hausman2018learning,shu2017hierarchical}, fine-tuned~\citep{frans2017meta,li2019hierarchical,li2019sub}, or used to guide the training of new related skills from scratch~\citep{igl2020multitask,tirumala2019exploiting,fox2016principled}. Crucially, the option set is not pruned, or selected from a larger set, in any manner, and the hierarchical policy needs to explore the combinatorial space of all options \& orderings to solve the new task. Option encoders~\citep{manoharan2020option} learn a small basis set embedding for a pretrained set of options; for a new task, they select among these bases and create new options using those bases. They do not address the retrieval problem directly.\\
\textbf{Context based Meta RL methods} try to infer task information from experience (multiple trajectories)~\citep{humplik2019meta, rakelly2019efficient}. They use a separate task encoder to predict task embeddings from trajectories which augments the state space of flat RL policies for quick generalization. The task encoder is trained using privileged information available during training. We also use similar privileged information (optimal trajectories returned by the {\sc HRLSolver}) for training our QGN and option embeddings but focus on hierarchical learning instead. Another related work by \cite{Sohn2020Meta} utilizes multiple trajectories to infer the \textit{subtask graph} that encodes subtask dependencies. The inferred graph is used by a graph execution policy to maximize reward by completing high reward tasks in the environment. In contrast, we do not require multiple trajectories at test time to index options. Also, we focus on efficient reuse of available options to accomplish a specific goal in the MDP instead of maximizing reward by accomplishing whatever subtasks can be completed in the environment.

\section{\oihrl: Learning an index over pretrained options}
\label{sec:methods}
\subsection{Preliminaries}
\label{sec:preliminaries}

\begin{figure*}[t]
    \begin{subfigure}[b]{0.35\textwidth}
      \centering
      \includegraphics[width=\textwidth]{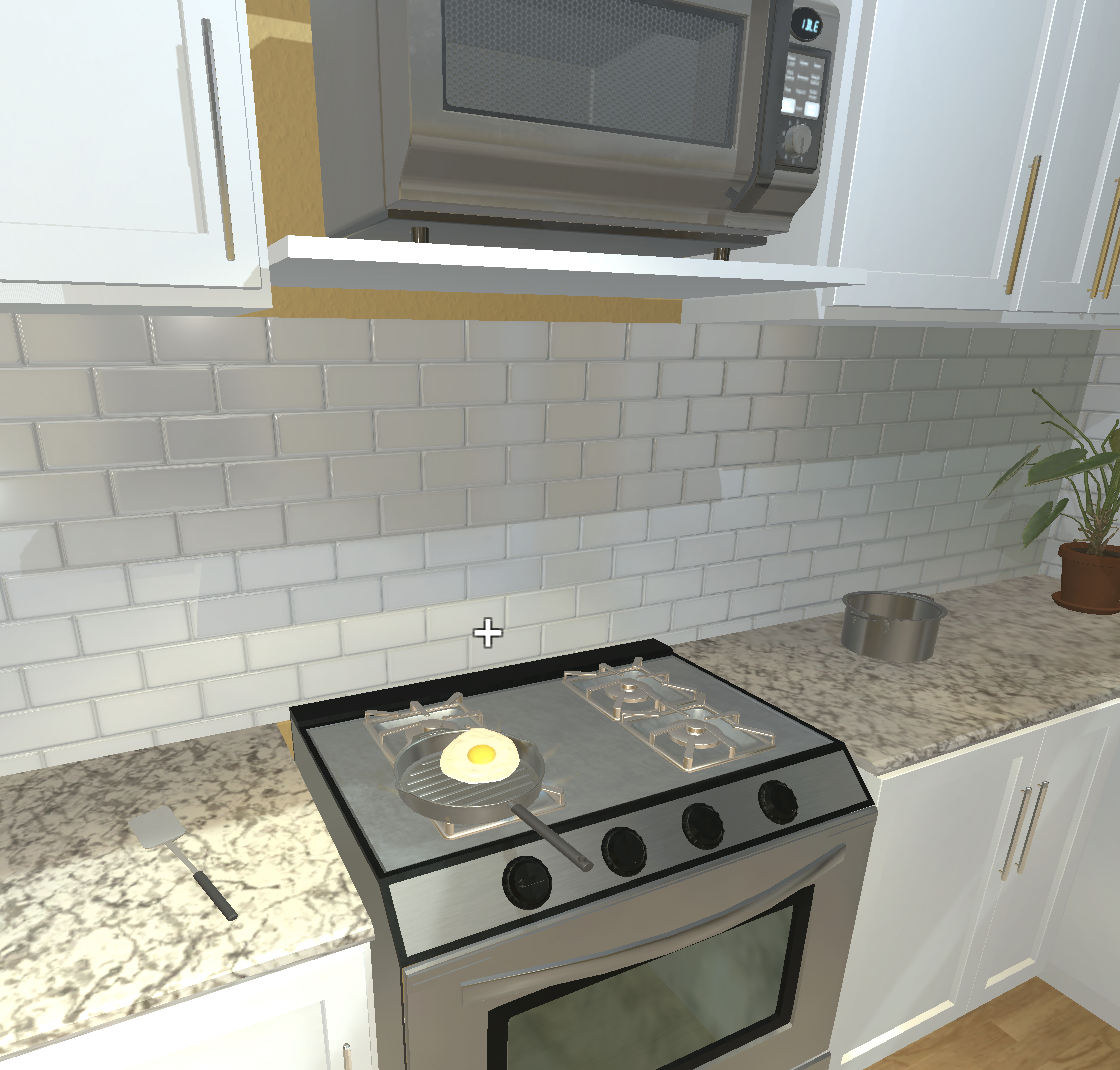}
      \caption{}
      
    \end{subfigure}
    \hfill
    \begin{subfigure}[b]{0.64\textwidth}
      \centering
      \includegraphics[width=\textwidth]{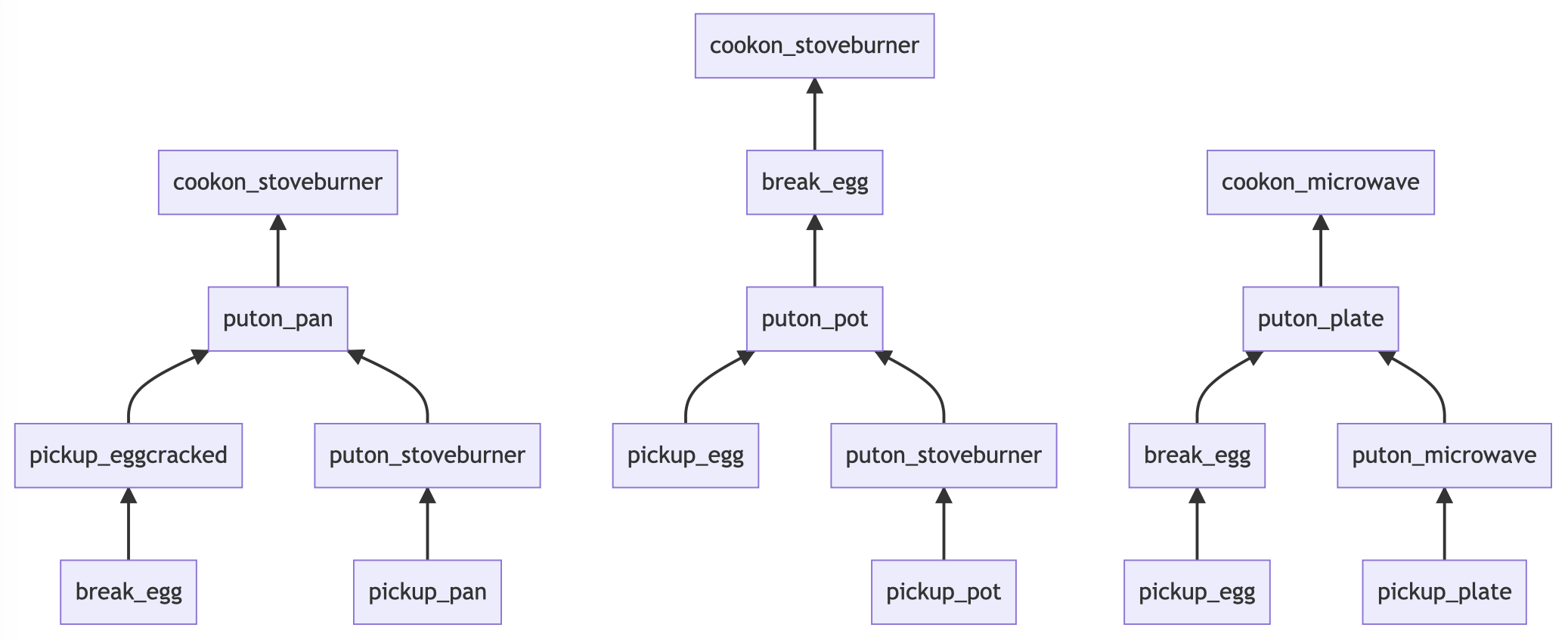}
      \caption{}
      \label{fig:omlette-recipes}
    \end{subfigure}
    \caption{Different ways to make an omelette in the AI2THOR kitchen. The task is making an omelette and three variants of it are shown.}
\end{figure*}

Consider a kitchen with various food items, utensils and appliances. An agent can accomplish various goals in this kitchen ranging from picking up a bowl, picking up an egg, picking up a pan, cracking an egg, cooking on a stove, making an omelette, slicing apples, placing on a table, etc. Notice that some goals, such as making an omelette, depend upon other, earlier goals having been achieved first, in a specific order. In this paper, we use the term \textit{task}, that an agent can perform, to refer to the accomplishment of a specific \textit{end goal}; thus, a task is said to be completed when its associated goal has been achieved. The conditions that need to be satisfied before a particular goal can be achieved (e.g. to make an omelette on a stove, we need to have a cracked egg and a lit stove) are referred to as \textit{preconditions} of the task. Also, each task can be completed in multiple ways. In Figure~\ref{fig:omlette-recipes} for example, the task of making an omelette can be completed by cooking on a stove or in a microwave. We call each such way to complete a task a \textit{task variant}. 

In general, we denote a task variant as $t_{i,j}$ where $i\in\{1,2,\ldots n\},\ j\in\{1,2,\ldots m_i\}$ and its preconditions as $\mcal{C}(t_{i,j}) \subset \{t_{i,j} | i=1\ldots n,\ j=1\ldots m_i\}$. A task $t_i = \{t_{i,1}, t_{i,2}, \ldots t_{i,{m_i}} \}$ is defined as the set of all task variants with the same goal.  Here $n$ is the number of possible tasks and $m_i$ is the number of variants of task $t_i$. Finally, the set of tasks is denoted using $\mcal{T} = \{t_1,t_2,\ldots t_n\}$.

When an environment is initialized, a goal task $t_{i}$ is chosen from $\mcal{T}$, and a particular task variant, say $t_{i, j}\ (j\in\{1,2,\ldots m_i\})$ is chosen. 
This chosen goal task $t_{i}$ is used to get a MDP $\mcal{M}_{t_i} = (\mcal{S}, \mcal{A}, \mcal{R}_{t_i}, \mcal{P})$ where $\mcal{S}$ and $\mcal{A}$ are the set of states and actions, $\mcal{R}_{t_i}$ and $\mcal{P}$ are the reward and transition functions. Here the reward function is sparse with the reward being non-zero only when the task $t_i$ is completed. 

Some tasks in the environment are simple and do not have any dependencies ($|\mcal{C}(t_{i, j})| = 0$). We call such tasks \textit{\primitives}. By design, base tasks only have a single variant ($|\mcal{C}(t_{i, j})| = 0  \implies m_i = 1$). We denote the set of base tasks $\{t_{1,1}, t_{2,1}, \ldots t_{k,1}\}$ as $\mcal{B}$ where $|\mcal{B}| = k$. A task in $\mcal{T}$ that is not a base task is called a \textit{composite} task.

We assume that there is a set of pretrained option policies $\mcal{O} = \{\opt_1,\opt_2, \ldots \opt_{k}\}$ for completing each base task. Given the \primitives, we define a task variant $t_{i,j}$'s \textit{recipe} recursively as the union of the recipes of all task variants in $\mcal{C}(t_{i, j})$, where the recipe of base tasks $t_{b,1} \in \mcal{B}$ is explicitly set to $\{t_{b,1}\}$. The \textit{recipe length} of a task variant $t_{i,j}$ is defined as the cardinality of its recipe.

\subsection{\oihrl\ objective}

Suppose an agent encounters a new task in a new environment. In the absence of prior experience, the agent will need to explore the environment and interact with it using primitive actions, as it learns to navigate the environment and accomplish various goals. For the agent to effectively use hierarchical reinforcement learning, it will then have to \textit{partition} the overall task into appropriate subtasks, and learn both how to accomplish the subtasks, as well as how to sequence them together in an appropriate manner. This is in general a very expensive process, due to the combinatiorial nature of the space of subtasks.

Instead, we ask the question: Suppose  the agent has significant experience with a range of related environments and tasks; how can it leverage its prior experience to accomplish the task on hand in an efficient manner? To build this novel lifelong-learning capability, we propose a two-step process: the agent first identifies which of a large set of prelearned options are relevant given the environment and its items, and then constructs a hierarchical policy using only the relevant options.

To solve this option retrieval problem, we propose to learn an \textit{affinity} or similarity measure between options and the initial state of the environment. This mechanism consists of an index $\mcal{I}$ and a Query Generation Network (QGN) $\mcal{N}$. %
The index $\mcal{I} \subset \R^d \times \mcal{O}$ stores the options $\mcal{O}$ together with a learnt key $\e(t_i) \in \R^d$ for option $\opt_i$, i.e. $\mcal{I} = \{(\e(t_i), \opt_i)\}$. The QGN generates a query $\mb q \in \R^d$ based on the objects initially present in the environment ($s_0$). The subset retrieved consists of the top-p \citep{holtzman2020nucleussampling} options based on $\hat{\mb p}_i = \operatorname{Softmax} ({\mb q}^{\intercal} \e(t_i))$ as shown in Algorithm~\ref{alg:retrieval}.

\begin{algorithm}[H]
  \caption{Querying the Index}
  \begin{algorithmic}[1]
  \State \textbf{Inputs:} Initial environment state $s_0$, Index $\mcal{I} = \{(\e(t_i), \opt_i)\}$, QGN $\mcal{N}$
  \Function{selectOptions}{$s_0$, $\mcal{I}$, $\mcal{N}$}
      \State query $\mb q \gets \mcal{N}(s_0)$
      \State similarities $\mb s \gets [{\mb q}^{\intercal} \e(t_1), {\mb q}^{\intercal} \e(t_2), \dots]$
      \State retrieval probabilities $\hat{\mb p} \gets \operatorname{Softmax}(\mb s)$
      \State retrieved option set $\widehat{\mcal{O}}$  s.t. $\widehat{\mcal{O}} \subset \mcal{O}$ is the smallest set that satisfies
      $\sum_{\opt_i \in \widehat{\mcal{O}}}  \hat{\mb p}_i > p$ (top-p fetch, $p$ is set to 0.9)
      \State \Return $\widehat{\mcal{O}}$, $\hat{\mb p}$
    \EndFunction
  \end{algorithmic}
  \label{alg:retrieval}
\end{algorithm}

\subsection{The \oihrl\ meta-training loop}

The QGN $\mcal{N}$ and option embeddings $\mb e(t_i)$ are meta-learned while the HRL policies are learned for each specific task. %
The psuedocode for training $\mcal{N}$ and $\mb e$ is presented in Algorithm~\ref{algo:train}.
In the training loop, first an MDP $\mcal{M}_{t_{i}}$ is sampled based on a sampled task variant $t_{i, j}$. Next a subset of options $\widehat{\mcal{O}}$ are selected using \textsc{selectOptions}.
This function also returns retrieval probabilities $\hat{\mb p}$ for all options. An HRL policy is then learned over options $\widehat{\mcal{O}}$. The policy is used to get a set of trajectories $\Lambda$\ %
which succeeded in completing the task. Next, $\mb y$, the normalized counts of the number of times each option was used by the HRL policy in $\Lambda$ is computed using \textsc{targetFromTrajectories}. Finally, $(\mcal{N}, \mathbf{e})$ are updated in order to minimize cross entropy error between $\hat{\mb p}$ and the normalized counts $\mb y$ as (recall $k = |\mcal{O}|$):

$$
\mcal{L}(\mb y, \hat{\mb p}) = - \frac{1}{k} \sum_{i=1}^{k} \mb y_{i}\log(\hat{\mb p}_i)
$$

\begin{figure}[ht]
\begin{minipage}[t]{0.5\textwidth}
\begin{algorithm}[H]\small
  \lineskiplimit=-\maxdimen
  \caption{Training}
  \begin{algorithmic}[1]
  \For{a number of train iterations}
    \State Sample train task variant $t_{i, j}$ and MDP $\mcal{M}_{t_i}$
    \State $\widehat{\mcal{O}}$, $\hat{\mb p}$ $\gets$ \textsc{selectOptions}($s_0$, $\mcal{I}$, $\mcal{N}$) %
    \State $\Lambda$ $\gets$ \textsc{HRLSolver}($\widehat{\mcal{O}}$, $\mcal{M}_{t_{i, j}}$)
    \State $\mb y$ $\gets$ \textsc{targetFromTrajectories}($\Lambda$)%
    \State Train $\mcal{N}$ and \textbf{e} to minimize $\mcal{L}(\mb y, \hat{\mb p})$
    \EndFor
  \end{algorithmic}
  \label{algo:train}
\end{algorithm}
\end{minipage}
\hfill
\begin{minipage}[t]{0.49\textwidth}
\begin{algorithm}[H]
  \lineskiplimit=-\maxdimen
  \caption{Testing}
  \begin{algorithmic}[1]\small
  \vspace{1.1em}
  \For{a number of test task variants}
    \State Sample task $t_{i,j}$ and MDP $\mcal{M}_{t_i}$
        \State $\widehat{\mcal{O}}$, $\hat{\mb p}$ $\gets$ \textsc{selectOptions}($s_0$, $\mcal{I}$, $\mcal{N}$)
        \State Train HRL policy using $\widehat{\mcal{O}}$
        \Statex
        \vspace{1.75pt}
      \EndFor
  \end{algorithmic}
  \label{algo:test}
\end{algorithm}
\end{minipage}
\end{figure}

\begin{algorithm*}[ht]
  \caption{Target from Trajectories}
  \begin{algorithmic}[1]
    \Function{targetFromTrajectories}{$\Lambda$}
    \State Let $R$ = sum of rewards of all trajectories in $\Lambda$
    \State Init $\mb y = [0, 0, \ldots\ k \text{ times} ]$
    \For{$T_i  = (o_{i,1}, o_{i,2}, \ldots o_{i,n}, \ r_i) \in \Lambda$}
        \For{$o_{i, j} \in T_i$}
            \State $\mb y[o_{i, j}] = \mb y[o_{i, j}] + \frac{r_i/n}{R}$
        \EndFor
     \EndFor
    \State \Return $\mb y$
    \EndFunction
  \end{algorithmic}
  \label{alg:target}
\end{algorithm*}

The goal of the meta-training phase is not to train a HRL policy for any given task,  but rather to learn associations between specific environments and the options relevant for those environments. Hence in practice for ease of computation, we assume access to an HRL Oracle which simulates a trained HRL policy. The primary advantage of using an oracle at train time is the reduction of redundant computations, as the indexing algorithm will need to repeatedly call HRLSolver on a related set of options--as such, the QGN can also be trained without the use of the oracle, albeit at higher computational cost.  Note that at test time, there is no access to oracular HRL policies, and a regular HRL policy is learned based only on options fetched from the index. 

\subsection{Using \oihrl\ for accomplishing new tasks}

In order to apply \oihrl\ for solving a new task, we simply provide the representation of the environment to the trained QGN $\mcal{N}$, and retrieve a set of relevant options. We then learn a HRL policy using the retrieved options, using Adavantage Actor Critic method~\citep{a2c}. If the retrieved set of options is sufficient for accomplishing the task, the HRL policy will succeed in completing the goal; further, the fewer the extraneous options retrieved, the faster the HRL policy can be learned. In the lifelong learning scenario we describe here, the agent has access to a large prelearned library of options, an effective retrieval mechanism is critical for effectively learning HRL policies \& completing new tasks.

In our experiments for evaluating the efficacy of the QGN $\mcal{N}$, we sample a number of task variants, each corresponding to an MDP to test on. None of the task variants in this test set have been encountered during training. For each  MDP, as described above, we fetch relevant options, train an HRL policy, and evaluate the learned HRL policy as a measure of the usefulness of the QGN.

\subsection{Handling incomplete fetches}
It is possible that in some cases the options selected using the \textsc{selectOptions} function (Algorithm~\ref{alg:retrieval}) are not sufficient to complete the goal task variant. If this happens, there is no way that the HRL policy will be able to converge and complete the task. However, the fetched set is still likely to include meaningful or relevant options, and represents a potentially better starting point than learning a policy from scratch. To address incomplete fetches, we can have the HRL learner use the fetched set of options and also allow it to act directly on the environment using primitive actions (e.g., UP, DOWN, etc. in a grid world). Alternatively, transfer learning algorithms like A2T~\citep{a2t} can be adapted to address this scenario.

\section{Experiments}
\label{sec:results}

We demonstrate the efficacy of \oihrl\ on two environments. We first use a grid-world environment (CraftWorld~\citep{craftworld}), where an agent's job is to navigate around to pickup simple objects available in the environment and combine them at a ``workshop" to make complex objects. In order to stress-test \oihrl, we developed a synthetic data generation scheme where tasks and their interrelated task variants %
were sampled by following a carefully designed procedure (Section~\ref{sec:craft}). To further demonstrate the applicability of \oihrl\ in more complex and realistic settings, we look at AI2THOR~\citep{ai2thor}. AI2THOR is an interactive 3D environment where an agent can navigate around and interact with a variety of objects. We focus on the kitchen scene and test our method on a number of diverse but interrelated food preparation tasks (Section~\ref{sec:ai2thor}).

\subsection{Baselines}
In order to study the effect of selecting a subset of options, we compare \oihrl\ with the following baselines having varying number of options made available to the HRL learner.
\begin{itemize}
    \item \textbf{\textsc{HRL-N}}: HRL using the set of options that correspond to task variants in the test task variant's recipe. This is an oracular upper bound on performance for \oihrl.
    \item \textbf{\textsc{HRL-N+K}}: same as {\sc HRL-N} but with k extra (irrelevant) options.
    \item \textbf{\textsc{HRL-Full}}: Hierarchical reinforcement learning using the entire library of options $\mcal{O}$.
\end{itemize}
Note that none of the baselines have a meta-learning stage and the option selection for them uses privileged information about the actual recipe of the goal task variant, something that is not available to \oihrl.

\subsection{CraftWorld Environment}
\label{sec:craft}
\subsubsection{Setup: Tasks and preconditions}

We look at 501 base tasks ($|\mcal{B}| = 501$), 500 of which correspond to navigating to and picking up 500 simple objects and one that corresponds to navigating to a workshop to combine collected objects to make a complex object. We have a set of pre-trained options $\mcal{O}$ ($|\mcal{O}| = |\mcal{B}|$) which act on the environment to perform these base tasks. We define 30 composite tasks, each corresponding to making a distinct complex object ($|\mcal{T}| = |\mcal{B}| + 30 = 531$). To come up with variants for these composite tasks, we assume that each simple object in the environment is similar to specific other objects which can substitute for it in any given recipe for a task variant. We divide the 500 objects in the environment into 100 groups (each of size 5 containing similar objects). A recipe over these groups is called a \textit{schema}. For each of the 30 composite tasks, we randomly sample a schema, each requiring 3 simple objects and a workshop. To generate task variants, we ground each schema using actual simple objects, giving us $125 (= 5^3\cdot1)$ variants for each composite task. This gives us a total of 3750 (125*30) task variants that are randomly split into train and test sets in a 80:20 ratio giving a total of 3000 distinct train recipes and 750 test recipes. Note that by design no train recipe can be exactly same as any test recipe and vice versa.

Let us consider a simplified example for this setting. Suppose we have 9 simple objects divided into three groups as: $g_1 = \{p_{11}, p_{12}, p_{13}\},\ g_2 = \{p_{21}, p_{22}, p_{23}\},\ g_3 = \{p_{31}, p_{32}, p_{33}\}$. The task set consists of 11 tasks where $t_1$ to $t_9$ correspond to navigating to and picking up simple objects. And $t_{10}$ and $t_{11}$ correspond to creating complex objects $c_1$ and $c_2$, the schemas for which are $(g_1, g_2)$ and $(g_2, g_3)$ respectively. These schemas are used to define the task variants for $t_{10}$ (i.e making $c_1$) as $t_{10,1} = \{p_{11}, p_{21}\}, t_{10,2} = \{p_{11}, p_{22}\}, t_{10,3} = \{p_{11}, p_{23}\}$ and so on, similarly for $t_{11}$. Here, the tasks are $\mcal{T} = \{t_1, t_2, \ldots t_{11} \}$ with the preconditions for $t_1$ through $t_9$ being empty. In this example, we have omitted the workshop for simplicity. In the actual tasks, after collecting the simple objects in a task variant's recipe, the agent needs to use a workshop to create the complex object (which is the goal of the task variant). For example, the preconditions $t_{10,1}$ will be $\{p_{11}, p_{21}, w\}$ where $w$ is a base task corresponding to navigating to a workshop and using it.

\subsubsection{Environment States and Dynamics}
For each environment, we sample one task variant as the current goal. The environment is initialized with all required simple objects and workshops required to complete the chosen task variant (making a complex object). Additionally, to make the retrieval non-trivial and to mimic real-world scenarios where other irrelevant things might be present, we add 2 other simple objects to the environment, which we call distractors. 
We use a binary representation for the environment state, denoting the objects present in the grid and an inventory which keeps track of the simple objects collected and complex objects made by the agent. When the agent picks up a simple object, it is removed from the grid and added to the inventory. Similarly, when using the workshop to make a complex object, the simple objects in its recipe are consumed from the inventory and the newly created complex object added.
The agent receives a reward of +1 on completing the task (using any variant). Collecting unnecessary simple or complex objects is penalized with a reward/penalty of -0.3. The rewards are sparse and the agent only receives them on successful completion of the goal task variant and intermediate steps are not rewarded. We use a maximum episode length of 10.

\subsubsection{\oihrl\ Training Details}
The Query Generation Network is chosen to be a two layered network with ReLU activations, hidden dimension 100 and key dimension of 50. The QGN $\mcal{N}$ and the Index $\mcal{I}$ are trained for 30000 iterations with a batch size of 32 using the method described in Algorithm~\ref{algo:train}. Evaluation follows the scheme described in Algorithm~\ref{algo:test}. The HRL policy is trained using Advantage Actor Critic (A2C)~\citep{a2c}.

\subsubsection{Results}
\begin{figure*}[t]
    \centering
    \includegraphics[width=\textwidth]{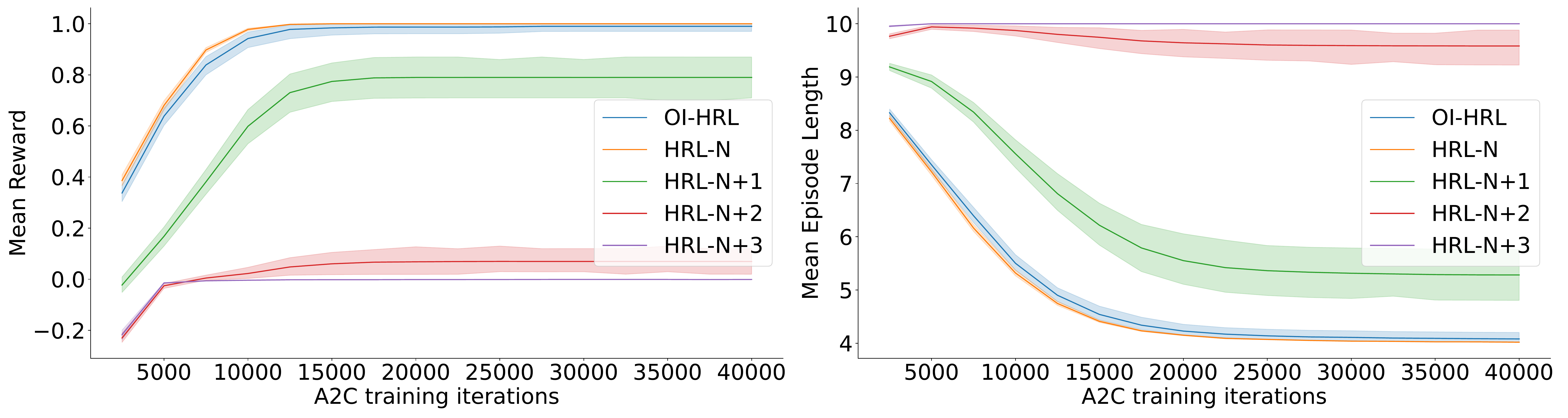}
    \caption{A2C policy training results using options fetched from \oihrl\ and other baselines on CraftWorld environment. Left: Mean reward vs A2C training iteration. Right: Mean Episode length vs A2C training iterations. HRL-N and \oihrl\ converge to near optimal policies with mean rewards close to 1 and episode lengths close to 4 while other baseline do not.}
    \label{fig:template}
\end{figure*}

\textbf{Task completion:} Figure~\ref{fig:template} shows the mean reward and mean episode length as a function of A2C training steps on the test set of tasks. The curves represent \oihrl\ and the various (noisy oracle) baselines discussed above, and are averaged over 100 test tasks with shaded regions showing 95\% confidence intervals. 

Since the baselines include all required options (HRL-N) along with some number $k$ of other irrelevant options (HRL-N+$k$), they represent aspirational targets for the performance of \oihrl; in particular, HRL-N is an oracular upper bound, while HRL-N+$k$ demonstrate varying degrees of imperfect-but-overcomplete fetches to show the importance of selecting a compact subset of options. 

\begin{figure}
    \centering
    \includegraphics[width=0.5\textwidth]{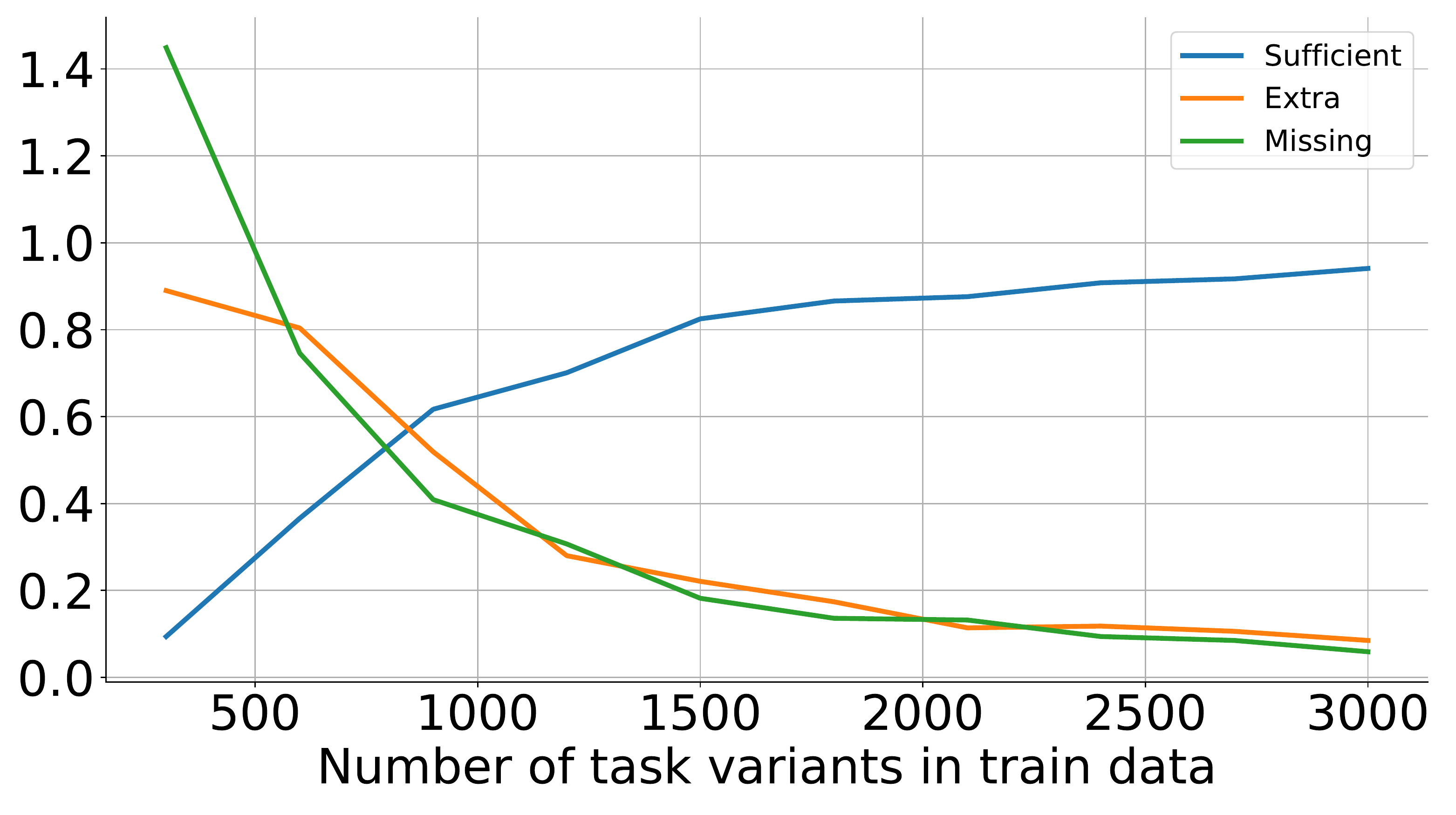}
    \caption{Effect of size of training data. The plot shows the fraction of test trials where a sufficient set of options was fetched (Sufficient), the average number of extra options fetched (Extra), and the number of missing options on average (Missing).}
    \label{fig:traindata}
\end{figure}%

First, we see that the oracular HRL-N always converges to the optimal policy with optimal mean reward of 1 and optimal mean episode length of 4, and that the variance across tasks is negligible. On adding just one extra option in HRL-N+1, the mean reward drops and mean episode length increases since the agent fails to find the optimal HRL policy for all tasks. Moving on to HRL-N+2, we see that only a small fraction of tasks are being completed resulting in near zero mean reward. On increasing the set of options available to HRL learner to the full set of 501\footnote{500 options to pick up objects and 1 to use the workshop} options (as is generally the case in prior work), there is no hope for any task to be completed successfully. \oihrl, on the other hand, fetches the relevant set of options based on the environment state making it much easier for the HRL learner to converge to an optimal policy. And as we can see, \oihrl\ achieves a high mean reward of about 0.99 which is better than HRL-N+1 but slightly lower than the upper bound of HRL-N.

Analysing the selected option sets, we find that on average, \oihrl\ fetches about 0.07 extra options and misses out on 0.01 options. While it is possible to learn when some extra options are fetched (as seen in HRL-N+1), when some required options are not fetched, there is no way in which the task can be completed.

\textbf{Effect of varying train data size on retrieval performance:} We now examine the value of \oihrl\ in a continual learning scenario; in particular, how does retrieval accuracy change as a function of the number of tasks already experienced by the agent? To measure this, we trained the QGN and index using varying amounts of train data and report the quality of the fetched options on 1000 test environments. Figure~\ref{fig:traindata} shows the fraction of test trials where sufficient options were fetched, along with the average number of options that were extra (irrelevant) or missing in the retrieved option set. We see that as more tasks are included in the training data, retrieval performance steadily improves, with sufficient options fetched on nearly all test tasks, and the number of extra options steadily dropping as well. This is because, with increased training data, the similarities / co-occurrences between groups of options in hierarchical plans are better captured in the QGN. %

\subsection{AI2THOR}
\label{sec:ai2thor}

The AI2THOR environment offers a more realistic testbed for \oihrl; in particular, it allows us to evaluate \oihrl\ on a significantly richer  distribution of tasks with (a) larger number of options on average per recipe, with higher variation across recipes, and (b) more complex option sequencing in accomplishing tasks.

\subsubsection{Setup: Tasks and preconditions}
We look at 43 base tasks ($|\mcal{B}| = 43$) that correspond to interacting with one of 21 possible objects in the environment. These base tasks are of types:
\begin{itemize}
    \item \textit{pickup\_$\langle$object$\rangle$} that involves navigating near the location of a desired object (e.g. egg, plate, bowl) and picking it up.
    \item \textit{puton\_$\langle$receptacle$\rangle$} that involves navigating near the location of a desired recepticle (e.g. dining table, pan, plate) and placing the object that is being held by the agent in/on it.
    \item \textit{cookon\_$\langle$appliance$\rangle$} that assumes that an object or recpetacle containing an object is already in/on a cooking appliance (a stoveburner or a microwave, for example) and involves turning the appliance on and off in order to cook it.
    \item \textit{slice\_$\langle$object$\rangle$} and \textit{break\_$\langle$object$\rangle$} that involves navigating to and slicing or breaking objects that are sliceable (a bread loaf, for example) or breakable (an egg, for example).
    \item \textit{fill\_$\langle$receptacle$\rangle$\_with\_$\langle$liquid$\rangle$} that involves filling certain receptacles like mug, cup or a bottle with one of three liquids - coffee, water or wine. 
\end{itemize}

We have a set of rule-based options $\mcal{O}$ ($|\mcal{O}| = |\mcal{B}|$) which act on the environment to perform these base tasks.

We come up with 48 diverse but interrelated composite tasks that, along with the base tasks, induce a densely connected task graph ($|\mcal{T}| = |\mcal{B}| + 48 = 91$). We define multiple variants for these composite tasks, with different recipes depending on the objects available in the environment. For example, an egg can be cooked using a stove or a microwave, depending on what is available (see Figure~\ref{fig:omlette-recipes} for example). Each of these task variants can be accomplished with a unique combination of (a subset of) the 43 options. This results in 1336 distinct \textit{task variants} across all 48 tasks with recipe length varying from 3 to 17. We use 812 task variants (60\%) for training, 262 (20\%) for validation and 262 (20\%) for testing. 

\subsubsection{Environment States and Dynamics}

We work with a structured state representation instead of visual features for simplicity. We use a 567 dimensional binary vector that encodes the presence/absence of objects in the scene and object metadata such as whether an object is currently picked up, whether it contains other object(s) or liquid in/on it, whether it is cooked, etc. To initialise the environment for a specific goal \textit{task variant}, we include the objects necessary for accomplishing the goal, omit objects that might cause alternative variants of the goal task to succeed to avoid ambiguity. Additionally, to make the retrieval non-trivial and generalisable, we introduce distractor objects by including each of the remaining objects with a 50\% probability.

For all options, we maintain a well defined set of preconditions and postconditions relating to the states of environment objects, that need to be satisfied before and after their execution. For example, the \textit{pickup\_plate} option requires the presence of a plate in the scene (precondition) and results in plate being picked up and its removal from the receptacle list of its parent receptacle (a dining table, for example). Since we are not concerned with the precise details of the agent’s 3D navigation and object manipulation as long as the preconditions and post-conditions are satisfied, we use a custom implementation of AI2THOR with SMDP state transition rules constructed from preconditions and postconditions of the known library of options $\mcal{O}$. As we are only concerned with executing ‘options’ and not low level ‘actions’ in the environment, we omit the implementation of the precise details of 3D world and the effects of actions. We assign a reward of +1 when the goal task is accomplished, and a step penalty of 0.002 with a maximum episode length of 500.

\subsubsection{\oihrl\ Training Details}

We use a two layered QGN with 10 hidden units and a key dimension of 10. The QGN $\mcal{N}$ and the index $\mcal{I}$ are trained for 100K iterations with a batch size of 32 using the method described in Algorithm~\ref{algo:train}. We perform hyperparameter search and select the best checkpoint using the validation tasks. Similar to the CraftWorld experiments, we use Algorithm~\ref{algo:test} to conduct evaluation. The HRL policy is trained using Advantage Actor Critic (A2C). 

\subsubsection{Results}
\begin{figure*}[htp]
\centering
\begin{subfigure}[b]{0.43\linewidth}
   \includegraphics[width=\textwidth]{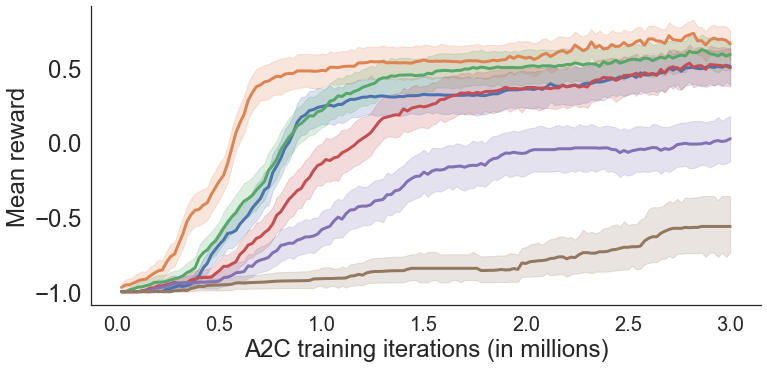}
   \caption{}
   \label{fig:ai2thor_reward}
\end{subfigure}
\hspace{\fill}
\begin{subfigure}[b]{0.55\linewidth}
  \includegraphics[width=\textwidth]{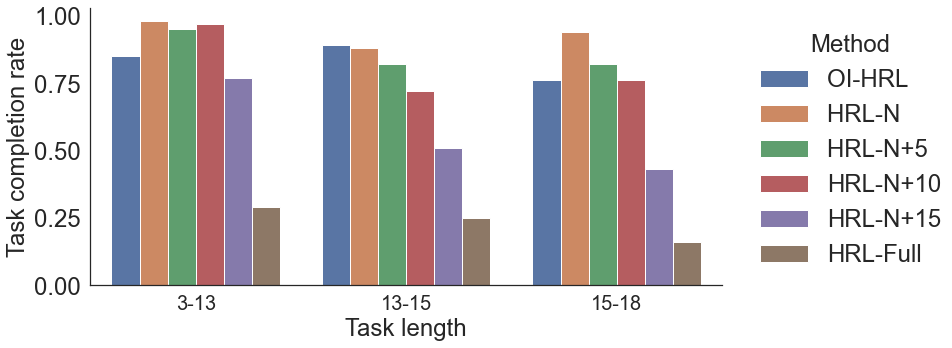}
  \caption{}
  \label{fig:ai2thor_task_len}
\end{subfigure}
\caption{A2C policy training results using options retrieved using \oihrl\ and other baselines \textbf{(a)} Mean reward vs A2C training iterations plot demonstrating the gains in sample efficiency. \textbf{(b)} Asymptotic task completion rates of the A2C policy for different task lengths.}
\end{figure*}

\textbf{Sample efficiency} Figure~\ref{fig:ai2thor_reward} shows the mean reward obtained in evaluation episodes by the A2C policy as a function of training steps (conventions are the same as in Figure~\ref{fig:template}). We train the policy for 3M iterations and report mean rewards averaged over 262 test tasks. As the test tasks are of widely varying difficulty, we see a much higher variance in mean rewards, even for the oracular HRL-N skyline. Addition of 5 extra options (HRL-N+5) sees only a small decrease in asymptotic performance (well within the shaded 95\% CI) but a noticeable decrease in sample efficiency for the first 1M iterations. Adding 10 extra options (HRL-N+10) also sees a noticeable drop in sample efficiency and a small drop in asymptotic performance. HRL-N+15 sees a significant drop in both metrics. The HRL-Full baseline is not able to attain positive rewards even after 3M iterations. \oihrl\ is able to match the sample efficiency of HRL-N+5 for the first 1M iterations and achieves a slightly a lower asymptotic performance, comparable to HRL-N+10. It is important to note that these baselines are oracular in nature and make use of privileged information, albeit with some added noise. It is always theoretically possible to complete the task variant and attain maximum reward using fetches from these baselines. Despite the absence of these guarantees, \oihrl\ remains competitive with these baselines.

\textbf{Task completion rate} We now look at the task completion rate in Figure~\ref{fig:ai2thor_task_len}, defined as the fraction of test task variants for with the A2C policy is able to find the solution when trained for 3M iterations. We perform this analysis for different recipe lengths - 3-13 (51 tasks), 13-15 (65 tasks), and 15-18 (146 tasks). Despite being the most diverse category, very few task variants fall in the 3-13 category due to skewed distribution of recipe lengths across 1336 task variants. This poses generalization difficulties for \oihrl\ retrieval, resulting in lower task completion rates as compared to HRL-N+5 and HRL-N+10. In the 13-15 recipe length category however, \oihrl\ performs on par with the HRL-N skyline despite the increased task difficulties. For tasks with recipe lengths greater than or equal to 15, \oihrl\ performs competitively with HRL-N+10. 

\section{Discussion}
\label{sec:discussion}

We presented \oihrl, an option indexing approach towards large-scale reuse of learned options in HRL, where only a small fraction of options are relevant to the task at hand. We proposed a method by which we can infer option relevance based on the state of the environment, and showed that these relevance or affinity scores can be effectively learned over a distribution of tasks in a meta-training framework.  Our results show an exponential improvement in task completion / average reward rates for \oihrl\ compared to baselines that include all available options, and relevant options plus some fixed number of irrelevant options. While we do not address the question of where the library of options come from, it is natural to expect a continual learning agent will acquire such a library of skills over the course of its lifetime. 

We are excited about the following directions for refining our \oihrl\ framework: (1) incorporating details about the goal (prerequisites, feature-based description etc) in the QGN, (2) including tasks encountered during training/testing also into the index, in a \textit{continual learning} framework, thereby growing the option library, (3) increasing test-time robustness by learning additional options when the retrieved option set is incomplete.  For the continual learning scenarios, a key objective is to minimize redundancy in the option index; this involves many interesting technical challenges.

\bibliography{reinflearn}
\bibliographystyle{collas2022_conference}

\newpage

\end{document}